%% file: cvpr2014.tex
\documentclass[10pt,twocolumn,letterpaper]{article}

\usepackage{cvpr}
\usepackage{times}
\usepackage{epsfig}
\usepackage{graphicx}
\usepackage{amsmath}
\usepackage{amssymb}
\usepackage{dsfont}
\usepackage{subfigure}
\usepackage{booktabs}
\usepackage{rotating}
\usepackage{multirow}
\usepackage{array}
\usepackage{grffile}
\usepackage{color}
\usepackage{tikz}

\definecolor{ForestGreen}{RGB}{40,166,40}
\newcommand*{\DashedArrow}[1][]{\mathbin{\tikz [baseline=-0.25ex,-latex, dashed,line width=1pt,#1] \draw [#1] (0pt,0.5ex) -- (1.0em,0.5ex);}}
\newcommand*{\Arrow}[1][]{\mathbin{\tikz [baseline=-0.25ex,-latex, solid,line width=1pt,#1] \draw [#1] (0pt,0.5ex) -- (1.0em,0.5ex);}}
\newcommand*{\ThickArrow}[1][]{\mathbin{\tikz [baseline=-0.25ex,-latex, solid,line width=2.2pt,#1] \draw [#1] (0pt,0.5ex) -- (1.0em,0.5ex);}}
\def\a{\alpha}

\def\lbd{\lambda}

\def\s{\sigma}

\def\p{\varphi}
\def\x2{\chi^2}

\def\1{\mathds{1}}

\graphicspath{{images/}}

\usepackage[pagebackref=true,breaklinks=true,letterpaper=true,colorlinks,bookmarks=false]{} %{hyperref}

\cvprfinalcopy % *** Uncomment this line for the final submission

\def\cvprPaperID{578} % *** Enter the CVPR Paper ID here

\ifcvprfinal\fi
\begin{document}

%%%%%%%%% TITLE
\title{Generalized Max Pooling}

\author{Naila Murray and Florent Perronnin\\
Computer Vision Group, Xerox Research Centre Europe
%{\tt\small Firstname.Lastname@xrce.xerox.com}
% For a paper whose authors are all at the same institution,
% omit the following lines up until the closing ``}''.
% Additional authors and addresses can be added with ``\and'',
% just like the second author.
% To save space, use either the email address or home page, not both
%\and
%Second Author\\
%Institution2\\
%First line of institution2 address\\
%{\tt\small secondauthor@i2.org}
}

\maketitle
%\thispagestyle{empty}

%%%%%%%%% ABSTRACT
\input{abstract}
\input{introduction}

\input{litreview}

\input{primal}

\input{dual}

\input{evaluation}
\input{conclusions}
\input{appendix}

{\small

\input{cvpr2014.bbl}
}

\end{document}

%% file: abstract.tex
\begin{abstract}
State-of-the-art patch-based image representations involve a pooling operation 
that aggregates statistics computed from local descriptors.
Standard pooling operations include sum- and max-pooling.
Sum-pooling lacks discriminability because 
the resulting representation is strongly influenced 
by frequent yet often uninformative descriptors, but only weakly 
influenced by rare yet potentially highly-informative ones.
Max-pooling equalizes the influence of frequent and rare descriptors
but is only applicable to representations that rely on 
count statistics, such as the bag-of-visual-words (BOV)
and its soft- and sparse-coding extensions.
We propose a novel pooling mechanism that 
achieves the same effect as max-pooling but is applicable beyond the BOV 
and especially to the state-of-the-art Fisher Vector
-- hence the name Generalized Max Pooling (GMP).
It involves equalizing the similarity between each patch and the pooled representation,
which is shown to be equivalent to re-weighting the per-patch statistics.
We show on five public image classification benchmarks that the proposed
GMP can lead to significant performance gains with respect to
heuristic alternatives.
\end{abstract}

%% file: introduction.tex
\section{Introduction}
\label{sec:intro}

This work is concerned with image classification. % using global image representations.
A state-of-the-art approach to representing an image from a collection of local descriptors consists of  
(i) encoding the descriptors using an embedding function $\p$
that maps the descriptors in a non-linear fashion into a higher-dimensional space; and 
(ii) aggregating the codes into a fixed-length vector using a pooling function.
Successful representations that fall within this framework include
the Bag-Of-Visual-words (BOV) \cite{sivic03bov,csurka04bov}, 
the Fisher Vector (FV) \cite{perronnin07fv}, 
the VLAD \cite{jegou2012aggregating}, 
the Super Vector (SV) \cite{zhou10sv} 
and the Efficient Match Kernel (EMK) \cite{bo2009efficient}.
In this work, we focus on step (ii): % (descriptor extraction and coding) 
the pooling step.

\begin{figure}[!ht]
\centering
\subfigure[Sum-pooling]{
    \includegraphics[width=0.45\linewidth]{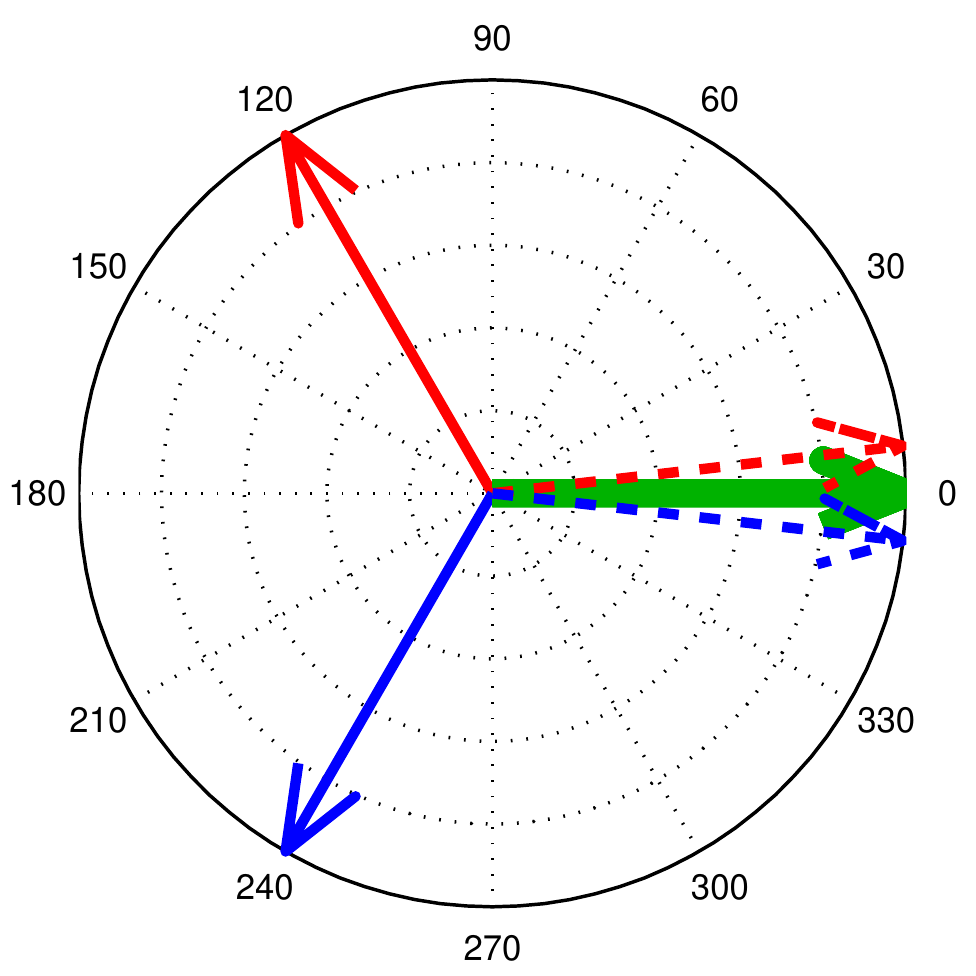}
    \label{fig:wgts_eg_sim}
}
\subfigure[Our GMP approach]{
    \includegraphics[width=0.45\linewidth]{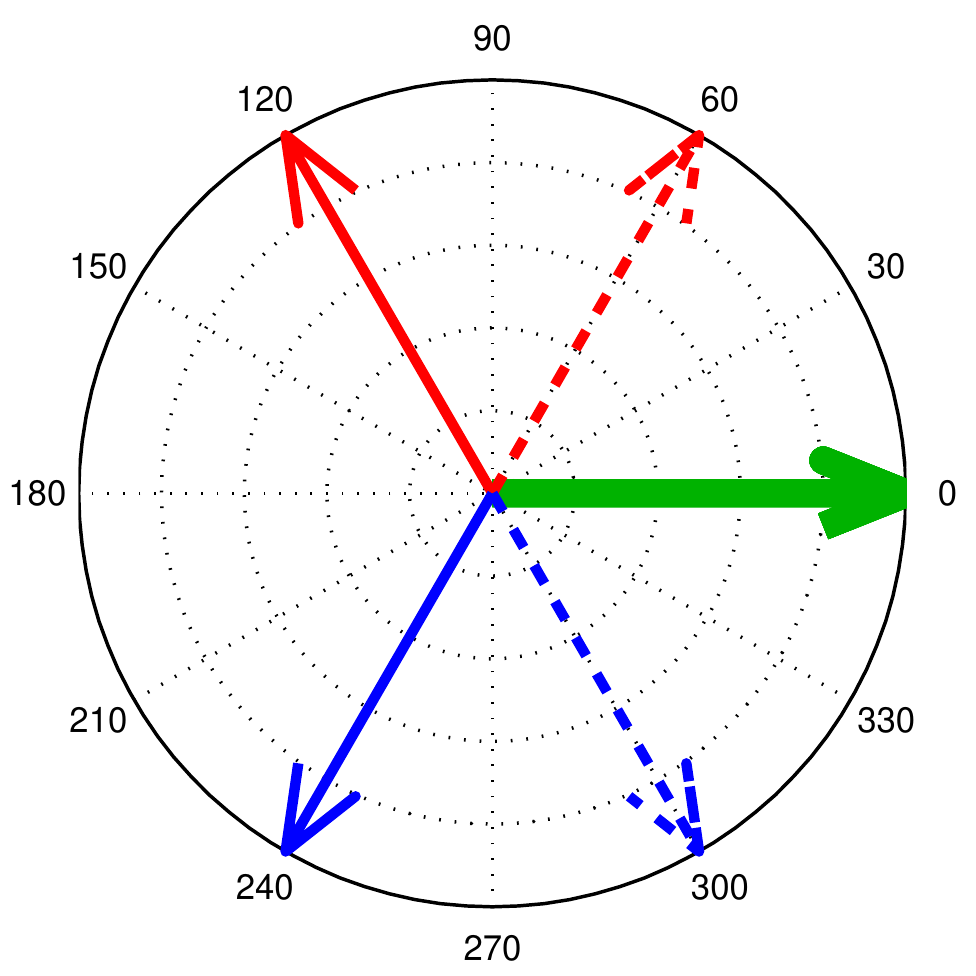}
    \label{fig:wgts_eg_diff}
}
\caption{We show the effect of pooling a single descriptor encoding (\textcolor{red}{$\Arrow[->]$} or \textcolor{blue}{$\Arrow[->]$})
with a set of tightly-clustered descriptor encodings (\textcolor{ForestGreen}{$\ThickArrow[->]$}).
Two pooled representations are shown: \textcolor{red}{$\Arrow[->]$} + \textcolor{ForestGreen}{$\ThickArrow[->]$} = \textcolor{red}{$\DashedArrow[->,densely dashed]$} 
and \textcolor{blue}{$\Arrow[->]$} + \textcolor{ForestGreen}{$\ThickArrow[->]$} = \textcolor{blue}{$\DashedArrow[->,densely dashed]$}. 
With sum-pooling (a), the cluster of descriptors \textcolor{ForestGreen}{$\ThickArrow[->]$} 
dominates the pooled representations \textcolor{red}{$\DashedArrow[->,densely dashed]$} 
and \textcolor{blue}{$\DashedArrow[->,densely dashed]$}, 
and as a result they are very similar to each other. 
With our GMP approach (b), both descriptors contribute meaningfully, 
resulting in highly distinguishable pooled representations.}
\label{fig:wgts_eg}
\end{figure}

By far the most popular pooling mechanism
involves summing (or averaging) the descriptor encodings \cite{csurka04bov,perronnin07fv,jegou2012aggregating,zhou10sv,bo2009efficient}.
%Given a set of local descriptors $\{x_1, \ldots, x_M\}$, 
%the sum-pooled representation is simply $\frac{1}{M}\sum_{i=1}^M \p(x_i)$.
An advantage of sum-pooling is its generality: it can be applied to any encoding.
A major disadvantage however is that frequently-occurring descriptors
%(sometimes referred to as bursty descriptors \cite{jegou2009burstiness})
will be more influential in the final representation than rarely-occurring ones (see Figure \ref{fig:wgts_eg_sim}).
By ``frequently-occurring descriptors" we mean descriptors which, though not necessarily identical, together form a mode in descriptor space. 
However, such frequent descriptors are not necessarily the most informative ones.
Let us take the example of a fine-grained classification task where the goal is to distinguish bird species.
In a typical bird image, most patches might correspond to background foliage or sky
and therefore carry little information about the bird class.
On the other hand, the most discriminative information might be highly localized 
and therefore correspond to only a handful of patches.
Hence, it is crucial to ensure that even those rare patches contribute significantly to the final representation.

Many approaches have been proposed in the computer vision literature to address the problem 
of frequent descriptors (see section \ref{sec:litreview}).
However, all of these solutions are heuristic in nature and/or limited to certain types of encodings. % and therefore not broadly applicable.
For instance, max-pooling \cite{boureau2010theoretical} 
only makes sense in the context of the BOV or its soft-coding and sparse-coding extensions.
However, it is not directly applicable to the FV, the VLAD, the SV or the EMK.
This is because the max-pooling operation treats each dimension independently while
for such representations the encoding dimensions are strongly correlated and should be treated jointly.

In this work we propose a novel pooling mechanism that involves equalizing the 
similarity between each patch and the pooled representation. 
This can be viewed as a generalization of max-pooling to any encoding $\p$ --
hence the name, {\em Generalized Max Pooling} (GMP).
%contribution of each patch when matched with the pooled representation.
%{\bf In this work we propose a novel pooling mechanism which consists in 
%{\em re-weighting} the descriptor embeddings.
%The weights are computed on a per-image basis to equalize the influence 
%of frequent and rare embedded descriptors (see Figure \ref{fig:wgts_eg_diff}).
%If $w_i$ denotes the weight associated with descriptor $x_i$,
%the weighted representation is $\sum_{i=1}^M w_i \p(x_i)$.}
%A major advantage of GMP is that it does not require quantizing descriptors 
%(as done for instance for the BOV) to detect frequent descriptors.
%As a consequence, it is general and can be applied in combination with any encoding function $\p$.
For instance, GMP is applicable to codebook-free representations such as the EMK and
to higher-order representations such as the FV. % \cite{chatfield11devil}.

Our main contributions are the following ones.
We first propose a matching criterion to compute the GMP representation (section \ref{sec:matching}).
It is referred to as the primal formulation because it involves explicitly computing the final pooled representation.
We then show that this criterion is equivalent to re-weighting the per-patch encodings before sum-pooling (section \ref{sec:wgh}).
We refer to this formulation as dual because the weights can be computed from the kernel of patch-to-patch similarities
without the need to access the patch encodings.
%We also provide an analysis of the proposed pooling mechanism and show that, 
%In the case of the BOV, it results in the well-known max-pooling (section \ref{sec:app}). 
%Hence, our approach can be viewed as a {\em Generalized Max Pooling} (GMP).
We show experimentally on five public benchmarks
that the proposed GMP can provide significant performance gains with respect to heuristic alternatives such as 
power normalization (section \ref{sec:exp}).

%% file: litreview.tex
\section{Related Work}\label{sec:litreview}

%Pooling in geometric space involves aggregating descriptors which satisfy some spatial constraints. 
%A standard geometric pooling approach involves using a spatial pyramid \cite{lazebnik2006beyond}, 
%where only descriptors extracted from the same cell of an image grid are aggregated. 
%In feature space, descriptors may be pooled within Voronoi regions (so-called ``hard" assignment) 
%or by their membership in one or more feature clusters (``soft" assignment). 
%\cite{boureau2010theoretical, boureau2011ask}.

The problem of reducing the influence of frequent descriptors 
has received a great deal of attention in computer vision.
This issue can be addressed at the pooling stage 
or \textit{a posteriori} by performing some normalization on the image-level pooled descriptor.
We therefore review related works on local descriptor pooling and image-level descriptor normalization.

\textbf{Local descriptor pooling.}
%\paragraph{Local descriptor pooling.}
Pooling is the operation which involves aggregating several local descriptor encodings into a single representation.
On the one hand, pooling achieves some invariance to perturbations of the descriptors.
On the other hand, it may lead to a loss of information.
To reduce as much as possible this loss, 
only close descriptors should be pooled together~\cite{boureau2010theoretical,boureau2011ask}, 
where the notion of closeness can be understood in the geometric and/or descriptor domains.
To enforce the pooling of close descriptors in the geometric space,
it is possible to use spatial pyramids \cite{lazebnik2006beyond}.
In the descriptor space, the closeness constraint is achieved through the choice of an appropriate embedding $\p$.
Note that our focus in this work is not on the choice of $\p$ 
but on the pooling operation, which we now discuss.

%An embedding $\p$ that relies on a visual vocabulary enforces 
%that only close descriptors in the descriptor space will be pooled together
%\cite{sivic03bov,csurka04bov,perronnin07fv,jegou2012aggregating,zhou10sv}.

%only descriptors assigned to the same image region are pooled together.
%Descriptor pooling may occur in geometric space or feature space. 
%Our focus in this work is on pooling in the descriptor space.

Pooling is typically achieved by
either summing/averaging or by taking the maximum response. % over the descriptor embeddings.
Sum-pooling has been used in many biologically-inspired visual recognition systems 
to approximate the operation of receptive fields in early stages of the visual cortex 
\cite{fukushima1982neocognitron,pinto2008real,jia2012beyond}.
It is also a standard component in convolutional neural networks \cite{le1989handwritten}.
A major disadvantage of sum-pooling is that it is based on the incorrect assumption
that the descriptors in an image are independent 
and that their contributions can be summed 
\cite{boureau2010theoretical,cinbis2012,sanchez2013image}.

Max-pooling was advocated by Riesenhuber and Poggio 
as a more appropriate pooling mechanism for higher-level visual processing 
such as object recognition \cite{riesenhuber1999hierarchical}. 
It has subsequently been used in computer vision models of object recognition \cite{serre2005object} 
and especially in neural networks \cite{ranzato2007sparse,lee2009convolutional}.
It has also recently found success in image classification tasks when used in conjunction with sparse coding techniques 
\cite{yang2009linear,boureau2010learning,%gao2010local,wang2010locality,
boureau2010theoretical}.
A major disadvantage of max-pooling is that it only makes sense when applied to 
embedding functions $\p$ which encode a strength of association
between a descriptor and a codeword, as is the case of the BOV
and its soft- and sparse-coding extensions. %***or sparse codes***.
However, it is not directly applicable to those representations which compute higher-order statistics 
such as the FV, which has been shown to yield state-of-the-art results in 
classification \cite{chatfield11devil} and retrieval \cite{jegou2012aggregating}.

%Average pooling in geometric space 
%used to achieve some translation invariance \cite{}. 
%Average pooling underlies standard patch descriptors such as the SIFT and HOG (see for instance \cite{bo2010kernel}).
%Average pooling in the descriptor space underlies image representations such as 
%the BOV \cite{sivic03bov,csurka04bov}, 
%the VLAD \cite{jegou2012aggregating}, 
%the Super Vector (SV) \cite{zhou10sv} 
%and the Efficient Match Kernel (EMK) \cite{bo2009efficient}.
%For such representations, 
%both geometric and descriptor pooling can be obtained using Spatial Pyramids \etal \cite{lazebnik2006beyond}
%which is equivalent to using an extended embedded function $\p$ (see for instance section 2 in \cite{bo2009efficient}).
%Since our focus is not on designing embedding function $\p$ but
%on the pooling mechanism, in what follows, 
%%, which may be constructed as summations of binary vectors whose non-zero elements indicate assignment to the corresponding visual word.

Several extensions to the standard sum- and max-pooling frameworks have been proposed.
For instance, one can transition smoothly from sum- to max-pooling using $\ell_p$- or softmax-pooling \cite{boureau2010theoretical}.
%It is also possible to incorporate class-dependent information in the geometric pooling mechanism \cite{feng2011geometric}. 
%One can also use more localized pooling to improve classification with small vocabularies \cite{boureau2011ask}.
It is also possible to add weights to obtain a weighted pooling \cite{deCampos2012}.
While our GMP can be viewed as an instance of weighted pooling (see section \ref{sec:wgh}),
a major difference with \cite{deCampos2012} is that their weights are computed using external information 
to cancel-out the influence of irrelevant descriptors while our weights are computed from the descriptors themselves and 
equalize the influence of frequent and rare descriptors.

%Pooling is performed to increase invariance within local regions while affording discriminability with respect to surrounding regions. An added motivation may be increasing the compactness of a given representation. A major challenge in constructing aggregation-based representations lies in avoiding the loss of critical information when performing averaging or max operations. This issue has typically been addressed on one of two fronts: (i) improving pooling regions so that only truly similar descriptors are pooled; (ii) reducing the influence of bursty descriptors to ensure a meaning contribution of all descriptors to an aggregated representation.

%\textbf{Improving pooling regions} 
%Feng \textit{et al.} proposed geometric $l_p$ pooling, which performs class-dependent geometric pooling 
%and as such better captures systematic geometric patterns in classes \cite{feng2011geometric}. 
%However, it is limited to categorization tasks. 
%Boureau \textit{et al.} proposed more localized pooling by creating bins within the Voronoi cells of K-means clusters \cite{boureau2011ask}, 
%as an alternative to larger vocabularies.

\textbf{Image-level descriptor normalization.} 
%\paragraph{Image-level descriptor normalization.} 
%While many feature pooling methods assume that local features are independent and identically distributed, this assumption is typically far from valid, particularly when dense sampling is performed \cite{boureau2010theoretical,cinbis2012}. Instead, particularly for repetitive structures such as building façades, descriptors often occur in bursts and, as mentioned in the introduction, can corrupt similarity measures between images. This is particularly problematic in fine-grained classification and recognition tasks and in instance-level retrieval tasks, where distinguishing characteristics between classes may be infrequent and contribute little to the aggregated representation. This problem is well-known in the text analysis domain, and is classically addressed using some form of term-frequency-inverse-document-frequency (TF-IDF) weighting.
%Recently, bursty visual descriptors have described and solutions proposed. 
Many works make use of sum-pooling and correct \textit{a posteriori} for the incorrect independence assumption
through normalization of the pooled representation.
J\'egou \textit{et al.} proposed several re-weighting strategies for addressing 
visual burstiness in the context of image retrieval by matching \cite{jegou2009burstiness}. 
These include penalizing multiple matches between a query descriptor and a database image 
and penalizing the matches of descriptors which are matched to multiple database images 
(\ie IDF weighting applied at the descriptor level rather than the visual word level). 
Torii \textit{et al.} proposed another re-weighting scheme for BOV-based representations, 
which soft-assigns to fewer visual words those descriptors which are extracted from a repetitive
image structure \cite{torii2013visual}. 
Arandjelovi\'{c} \textit{et al.} showed that, for the VLAD representation \cite{jegou2012aggregating}, 
applying $\ell_2$-normalization to the aggregated representation of each pooling region mitigates the burstiness effect \cite{arandjelovicall}. 
Delhumeau \etal found that, for VLAD, $\ell_2$-normalizing the descriptor residuals and then applying PCA before pooling was beneficial 
\cite{delhumeau2013revisiting}. 
Power normalization has also been shown to be an effective heuristic for treating 
frequent descriptors in BOV, FV or VLAD representations \cite{perronnin2010explicit,perronnin2010improving,jegou2012aggregating}.
The main drawback of the previous works  is that they are heuristic and/or restricted to image representations based on a finite vocabulary. 
In the latter case, they are not applicable to codebook-free representations such as the EMK.

One of the rare works which considered the independence problem in a principled manner
is that of Cinbis \etal which proposes a latent model to take into account inter-descriptor dependencies \cite{cinbis2012}.
However, this work is specific to representations based on Gaussian Mixture Models.
In contrast, our GMP is generic and applicable to all aggregation-based representations.

%% file: primal.tex
\section{GMP as equalization of similarities}
\label{sec:matching}

%\section{Primal Formulation}
%\label{sec:primal}

%\subsection{BOV intuition}

Let $X = \{x_1, \ldots, x_N\}$ be a set of $N$ patches extracted from an image
and let $\p_n=\p(x_n)$ denote the $D$-dimensional encoding of the $n$-th patch.

Our goal is to propose a pooling mechanism that mimics the desirable properties of max-pooling in the BOV case
and is extensible beyond the BOV.
One such property is the fact that the dot-product similarity between the max-pooled
representation $\p^{max}$ and a single patch encoding $\p_n$ is a constant value\footnote{
As mentioned in the introduction, we consider in this work the image classification problem.
For efficiency purposes, we focus on the case where the pooled representations
are classified using linear kernel machines, for instance linear SVMs.
Therefore, the implicit metric which is used to measure the similarity between 
patch encodings is the dot-product.}.
To see this, let $C$ denote the codebook cardinality ($C=D$ in the BOV case)
and let $i_n$ be the index of the closest codeword to patch $x_n$.
$\p_n$ is a binary vector with a single non-zero entry at index $i_n$.
$\p^{max}$ is a binary representation where a 1 is indicative of the presence of the codeword in the image.
Consequently, we have $\p_n^T \p^{max} = 1$ for all $\p_n$
and $\p^{max}$ is equally similar to frequent and rare patches.
%have an equal contribution in the matching.
This occurs because frequent and rare patches 
contribute equally to the aggregated representation, 

In contrast, the sum-pooled representation
is overly influenced by frequent descriptors.
Indeed, in the case of the unnormalized sum-pooled representation, we have
$\p^{sum} = [\pi_1, \ldots, \pi_C]^T$
where $\pi_k$ is the number of occurrences of codeword $k$ and therefore
$\p_n^T \p^{sum} = \pi_{i_n}$.
Consequently, more frequent patches make a greater contribution to the aggregated representation.
As a result, $\p^{sum}$ is more similar to frequent descriptors than to rare ones.

Figure~\ref{fig:wgts_eg} illustrates this property for our GMP approach: $\p^{sum}$ is more similar to the more frequent descriptor (Figure~\ref{fig:wgts_eg_sim}) whereas $\p^{gmp}$ is equally similar to both the frequent and rare descriptor (Figure~\ref{fig:wgts_eg_diff}), although the descriptors are not single-entry binary vectors as in the BOV case.
We now describe how GMP generalizes this property to arbitrary descriptors (section \ref{sec:primal}).
%We now generalize this property and introduce our GMP approach (section \ref{sec:primal}).
We also explain how to compute efficiently the GMP representation in practice (section \ref{sec:comp}).

%\subsection{Matching interpretation}
\subsection{Primal Formulation}
\label{sec:primal}

Let $\p^{gmp}$ denote the GMP representation.
We generalize the previous matching property and enforce
the dot-product similarity between each patch encoding $\p_n$ 
and the GMP representation $\p^{gmp}$ to be a constant $c$:
\begin{equation} \label{eqn:match}
\p_n^T \p^{gmp} = c, \mbox{ for } n=1,\ldots, N.
\end{equation}
Note that the value of the constant has no influence
as we typically $\ell_2$-normalize the final representation.
Therefore, we arbitrarily set this constant to $c=1$.
Let $\Phi$ denote the $D \times N$ matrix
that contains the patch encodings: $\Phi = [\p_1, ..., \p_N]$.
In matrix form, (\ref{eqn:match}) can be rewritten as:
\begin{equation} \label{eqn:primal}
\Phi^T \p^{gmp} = \1_N,
\end{equation}
where $\1_N$ denotes the $N$-dimensional vector of all ones.
This is a linear system of $N$ equations with $D$ unknowns.
In general, this system might not have a solution (\eg when $D<N$)
or might have an infinite number of solutions (\eg when $D>N$).
Therefore, we turn (\ref{eqn:primal}) into a least-squares regression problem and seek
\begin{equation} \label{eqn:primal_ls}
\p^{gmp} = \arg \min_\p ||\Phi^T \p - \1_N||^2,
\end{equation}
with the additional constraint that $\p^{gmp}$ has minimal norm in the case of an infinite number of solutions.
%We formulate problem (\ref{eqn:lineareq}) as a least-squares regression problem.
%and introduce a regularization term to avoid the problematic cases where the system of equations either does not
%admit any exact solution or admits multiple solutions. 
%We therefore solve:
%%
%\begin{equation} \label{eqn:primal}
%\p^* = \arg \min_\p \frac{1}{2M} \sum_{i=1}^M (\p_i^T \p - 1)^2 % + \frac{\lbd}{2} ||\p||^2
%\end{equation}
%%
The previous problem admits a simple closed-form solution:
%Then the minimum-norm solution to (\ref{eqn:primal}) has a simple closed form formula:
%
\begin{equation} \label{eqn:primalsol}
\p^{gmp} = (\Phi^T)^+ \1_N =  (\Phi \Phi^T)^+ \Phi \1_N,
\end{equation}
where $^+$ denotes the pseudo-inverse
and the second equality stems from the property $A^+=(A^TA)^+A^T$.
%by setting $A=\Phi^T$, we can rewrite equation (\ref{eqn:primalsol1}) as:
%%
%\begin{equation} \label{eqn:primalsol}
%\p^* = (\Phi \Phi^T)^+ \Phi \1_M
%\end{equation}
%%
We note that $\Phi \1_N = \sum_{n=1}^N \p(x_n) = \p^{sum}$ is the sum-pooled representation. 
Hence, the proposed GMP involves projecting 
the $\p^{sum}$ on $(\Phi \Phi^T)^+$.
Note that this is different from recent works in computer vision advocating for
the decorrelation of the data \cite{jegou2012negative}
since in our case the uncentered correlation matrix $\Phi \Phi^T$ is computed {\em from patches of the same image}.

It can be easily shown that, in the BOV case (hard coding), the GMP representation
is strictly equivalent to the max-pooling representation.
This is not a surprise given that GMP was designed to mimic the good properties of max-pooling.
We show in the appendix a more general property.

\subsection{Computing the GMP in practice}
\label{sec:comp}

Since the pseudo-inverse is not a continuous operation 
%(in the sense that if the sequence of matrices $A_n$ converges
%to $A$, then the sequence $A_n^+$ does not necessarily converge to $A^+$),
it is beneficial to add a regularization term to obtain a stable solution.
We introduce $\p^{gmp}_\lbd$, the regularized GMP:
\begin{equation} \label{eqn:primalreg}
\p^{gmp}_\lbd = \arg \min_\p ||\Phi^T \p - \1_N||^2  + \lbd ||\p||^2.
\end{equation}
This is a ridge regression problem whose solution is
\begin{equation} \label{eqn:primalsolreg}
\p^{gmp}_\lbd = (\Phi \Phi^T + \lbd I)^{-1} \Phi \1_N.
\end{equation}
The regularization parameter $\lbd$ should be cross-validated.
For $\lbd$ very large, we have $\p^{gmp}_\lbd \approx  \Phi \1_N / \lbd$
and we are back to sum pooling.
Therefore, $\lbd$ does not only play a regularization role.
It also enables one to smoothly transition between the solution to (\ref{eqn:primalsol}) ($\lbd=0$)
and sum pooling ($\lbd \rightarrow \infty$).

%\subsection{Computing $\p^{gmp}$}
We now turn to the problem of computing $\p^{gmp}_\lbd$.
We can compute (\ref{eqn:primalsolreg}) 
using Conjugate Gradient Descent (CGD)
which is designed for PSD matrices. % which may be sparse.
This might be computationally intensive if the encoding dimensionality $D$ 
is large and the matrix $\Phi$ is full (cost in $O(D^2)$).

However, we can exploit the structure of certain patch encodings.
Especially, the computation can be sped-up if the individual patch encodings $\p_n$ are block-sparse. 
By block-sparse we mean that 
the indices of the encoding can be partitioned into a set of groups where the activation
of one entry in a group means the activation of all entries in the group.
This is the case for instance of the VLAD and the SV where
each group of indices corresponds to a given cluster centroid.
This is also the case of the FV if we assume a hard assignment model
where each group corresponds to the gradients with respect to the parameters of a given Gaussian.
In such a case, the matrix $\Phi \Phi^T$ is block-diagonal. 
Consequently $\Phi \Phi^T + \lbd I$ is block diagonal
and (\ref{eqn:primalsolreg}) can be solved block-by-block,
which is significantly less demanding than solving the full problem directly (cost in $O(D^2/C)$).

%% file: dual.tex
\section{GMP as weighted pooling}
\label{sec:wgh}
%\section{Dual Formulation}
%\label{sec:dual}

In what follows, we first provide a dual interpretation of GMP as a weighted pooling (section \ref{sec:dual}).
We then visualize the computed weights (section \ref{sec:vis}).

%\subsection{GMP as weighted pooling}
%\label{sec:wgh}
\subsection{Dual Formulation}
\label{sec:dual}

%We now provide a dual interpretation of our GMP.
We note that the regularized GMP $\p^{gmp}_\lbd$ is the solution to (\ref{eqn:primalreg})
and that, consequently, according to the representer theorem, 
$\p^{gmp}_\lbd$ can be written as a linear combination of the encodings:
\begin{equation} \label{eqn:rep}
\p^{gmp}_{\lbd} = \Phi \a_{\lbd} 
\end{equation}
where $\a_\lbd$ is the vector of weights.
Therefore GMP can be viewed as an instance of weighted pooling \cite{deCampos2012}.
By introducing $\p=\Phi \a$ in the GMP objective (\ref{eqn:primalreg}),
we obtain: 
\begin{equation}
\a_{\lbd} = \arg \min_\a ||\Phi^T \Phi \a - \1_N||^2  + \lbd ||\Phi \a||^2.
\end{equation}
If we denote by $K=\Phi^T\Phi$ the $N \times N$ kernel matrix of patch-to-patch similarities,
we finally obtain:
\begin{equation} \label{eqn:dualreg}
\a_{\lbd} = \arg \min_\a ||K \a - \1_N||^2  + \lbd \a^T K \a
\end{equation}
which admits the following simple solution:
\begin{equation} \label{eqn:dualregsol}
\a_{\lbd} = (K+\lbd I_N)^{-1} \1_N 
\end{equation}
which only depends on the patch-to-patch similarity kernel,
not on the encodings.

We note that, in the general case, computing the kernel matrix $K$ and solving (\ref{eqn:dualreg}) 
have a cost in $O(N^2D)$ and $O(N^2)$ respectively.
This might be prohibitive for large values of $N$ and $D$.
However, as was the case for the primal formulation, 
we can exploit the structure of certain encodings.
This is the case of the VLAD or the FV (with hard assignment):
since the encoding is block-sparse, the matrix $K$ is block-diagonal.
Using an inverted file type of structure, one can reduce the cost of computing $K$ to $O(N^2D/C^2)$
by matching only the encodings $\p_n$ that correspond to patches assigned to the same codeword\footnote{The $O(N^2D/C^2)$ complexity is based on the optimistic assumption that the same number of patches $N/C$ is assigned to each codeword.}.
Also, one can solve for $\a_{\lbd}$ block-by-block which reduces the cost to $O(N^2/C)$.

Once weights have been computed, the GMP representation is obtained by linearly re-weighting the
per-patch encodings -- see equation \eqref{eqn:rep}.
Note that in all our experiments we use the primal formulation to compute $\p^{gmp}_{\lbd}$,
which is more efficient than first computing the weights and then re-weighting the encodings.
However, we will see in the following section that the dual formulation
is useful because it enables visualizing the effect of the GMP.

\subsection{Visualizing weights}
\label{sec:vis}

\begin{figure}[!h]
\centering
\begin{tabular}{ccc}
\vspace{-5mm}
\includegraphics[width=0.85\linewidth]{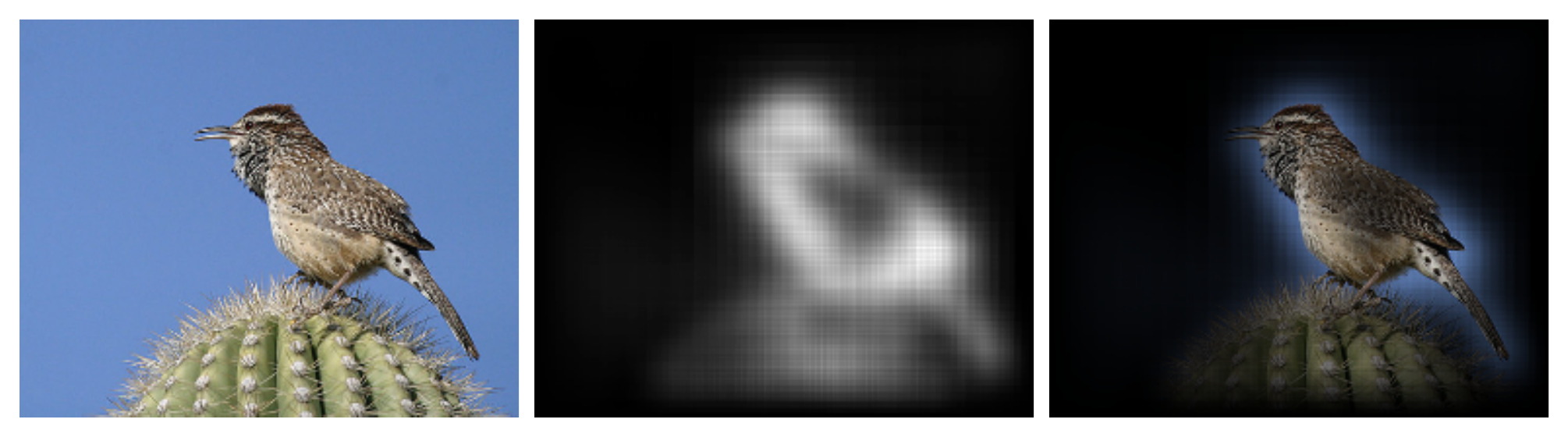} \\
\vspace{-5mm}
\includegraphics[width=0.85\linewidth]{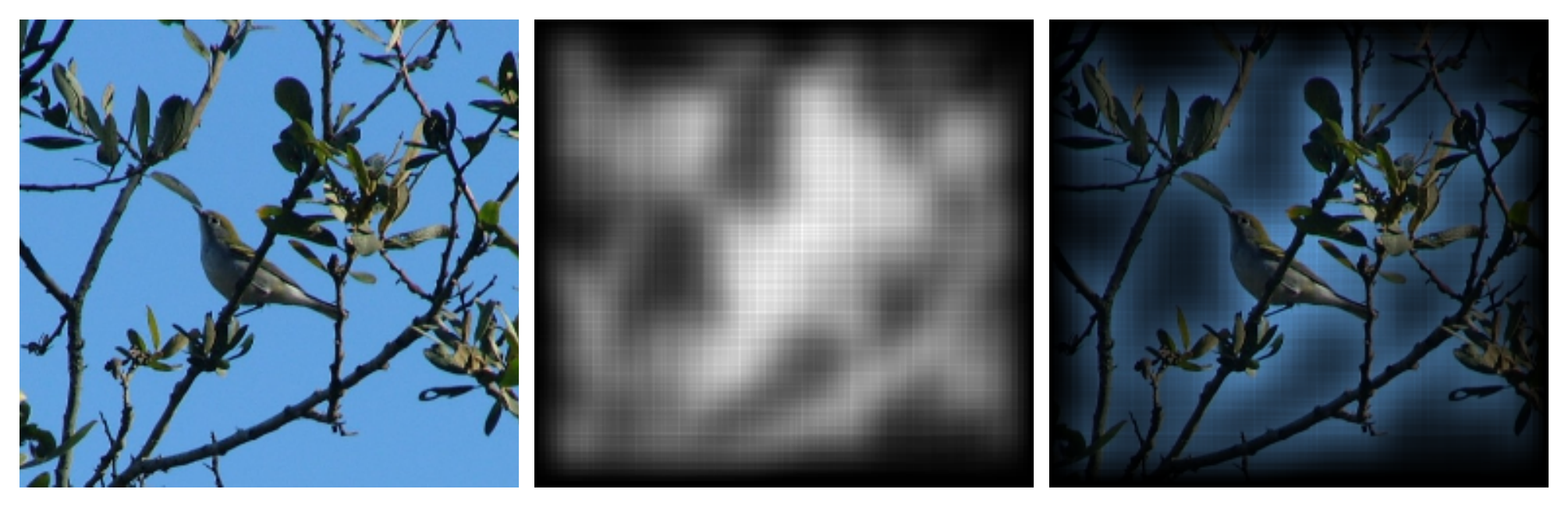} \\
\vspace{-0mm}
\includegraphics[width=0.85\linewidth]{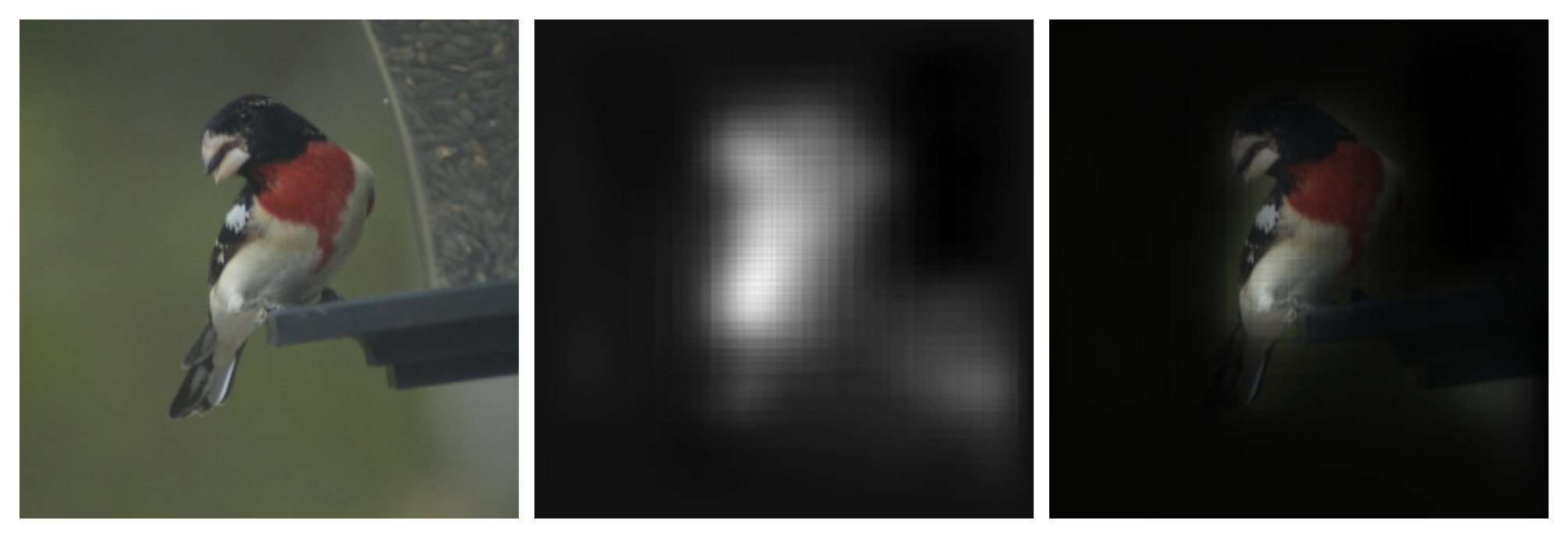} \\
\end{tabular}
\caption{Maps generated using weights computed from color descriptors using the EMK encoding,
for a sample of images from the CUB-2011 dataset.
Left: original images. Middle: weight maps. Right: weighted images.
%Note that these weighted images are only shown for illustrative purposes but are never used in practice.
See sections~\ref{sec:data} and \ref{sec:desc} for details of the dataset and descriptors.}
\label{fig:map_egs}
\end{figure}

We use weights computed via the dual formulation to generate a topographic map whose value at pixel location $l$ is computed as
\begin{equation}
s_l = \sum_{x_i \in \mathcal{P}_l} \a_i,
\end{equation}
where $\mathcal{P}_l$ is the set of patches $x_i$ which contain location $l$ and $\a_i$ is the weight of patch $x_i$. 
These maps give us a quantitative measure of the rarity, and potential discriminativeness, 
of the different regions in the image. 
Figure~\ref{fig:map_egs} shows several such maps for bird images. % in the case of the EMK encoding.
Clearly, they are reminiscent of saliency maps computed to predict fixations of the human gaze.
One sees that the highly-weighted regions contain rare image patches,
such as the breast of the bird in the third row. 
When the background is quite simple, as is the case of the first row,
the learned weights tend to segment the foreground from the background.
Note however that our maps are computed in a fully {\em unsupervised} manner and that there is no 
foreground/background segmentation guarantee in the general case.

%% file: evaluation.tex
\section{Experimental Evaluation}\label{sec:exp}

We first describe the image classification datasets % (section \ref{sec:data}) 
and image descriptors % (section \ref{sec:desc}) 
we use.
We then report results.

\subsection{Datasets} \label{sec:data}
%We benchmark the proposed GMP on five public datasets.
As mentioned earlier, we expect the proposed GMP to be particularly beneficial on fine-grained tasks
where the most discriminative information might be associated with a handful of patches.
Therefore, we validated the proposed approach on four fine-grained image classification datasets:
CUB-2010, CUB-2011, Oxford Pets and Oxford Flowers.
We also include the PASCAL VOC 2007 dataset in our experiments
since it is one of the most widely used benchmarks in the image classification literature.
On all datasets, we use the standard training/validation/test protocols.
\\[2mm]
The \textbf{Pascal VOC 2007} (VOC-2007) dataset \cite{pascal-voc-2007} contains 9,963 images of 20 classes.
Performance on this dataset is measured with mean average precision (mAP). 
A recent benchmark of encoding methods \cite{chatfield11devil} reported 61.7\% using the FV descriptor with spatial pyramids
~\cite{perronnin2010improving}. \\[2mm]
The \textbf{CalTech UCSD birds 2010} (CUB-2010) dataset \cite{WelinderEtal2010} contains 6,033 images of 200 bird categories.
Performance is measured with top-1 accuracy.
The best reported performance we are aware of for CUB-2010 is 17.5\%~\cite{angelova13}. 
This method uses sparse coding in combination with object detection and segmentation prior to classification. 
Without detection and segmentation, performance drops to 14.4\% \cite{angelova13}.\\[2mm]
The \textbf{CalTech UCSD birds 2011} (CUB-2011) dataset \cite{WahCUB_200_2011} is an extension of CUB-2010 that 
contains 11,788 images of the same 200 bird categories.
Performance is measured with top-1 accuracy.
The best reported performance for CUB-2011 is, to our knowledge, 56.8\%.
This was obtained using ground-truth bounding boxes and part detection \cite{berg2013}.
The best reported performance we are aware of that does not use ground-truth annotations 
or localization is 28.2\% \cite{Rodriguez2013}.\\[2mm]
The \textbf{Oxford-IIIT-Pet} (Pets) dataset \cite{parkhi12a} contains 7,349 images of 37 categories of cats and dogs. 
Performance is measured with top-1 accuracy. 
The best reported performance for Pets is 54.3\%, 
which was also obtained using the method of \cite{angelova13}. 
Without detection and segmentation,
performance drops to 50.8\% \cite{angelova13}.\\[2mm]
The \textbf{Oxford 102 Flowers} (Flowers) dataset \cite{Nilsback08} contains 8,189 images of 102 flower categories. 
Performance is measured with top-1 accuracy. 
The best reported performance for Flowers is 80.7\%, and was also obtained using the method of \cite{angelova13}. 
Again, without detection and segmentation performance drops to 76.7\% \cite{angelova13}.
%
%For each dataset, we use the standard evaluation metric:
%mAP on VOC 2007 and top-1 accuracy for the four fine-grained datasets.

\subsection{Descriptors} \label{sec:desc}

In our experiments, patches are extracted densely at multiple scales resulting in approximately 10K descriptors per image.
We experimented with two types of low-level descriptors:
128-dim SIFT descriptors \cite{lowe2004distinctive} and 96-dim color descriptors \cite{clinchant2007xrces}.
In both cases, we reduced their dimensionality to 64 dimensions with PCA.
Late fusion results were obtained by 
evaluating an unweighted summation of the scores given by the SIFT and color-based classifiers.

As mentioned earlier, the proposed GMP is general and can be applied to any aggregated representation. % obtained by max-pooling.
Having shown in the appendix, and verified experimentally, the formal equivalence between GMP and standard max-pooling in the BOV hard-coding case,
we do not report any result for the BOV.
In our experiments, we focus on two aggregated representations: 
the EMK~\cite{bo2009efficient} and the FV~\cite{perronnin07fv}.

\subsection{Results with the EMK}\label{sec:results_emk}

To compute the EMK representations we follow~\cite{bo2009efficient}:
we project the descriptors on random Gaussian directions, apply a cosine non-linearity
and aggregate the responses.
The EMK is a vocabulary-free approach which does not perform any quantization 
and as a result preserves minute and highly-localized image details. 
The EMK is thus especially relevant for fine-grained problems. % where such image details are necessary for good discrimination.
However, since all embeddings are pooled together rather than within Voronoi regions as with vocabulary-based approaches, 
the EMK is particularly susceptible to the effect of frequent descriptors.
Therefore we expect GMP to have a significant positive impact on the EMK performance.
To the best of our knowledge, there is no previous transformation that may be applied to the EMK to counteract frequent descriptors.
In particular, power normalization heuristics which are used for vocabulary-based approaches such
as the BOV~\cite{perronnin2010explicit} or the FV~\cite{perronnin2010improving} are not suitable.

The EMK representation has two parameters:
the number of output dimensions $D$ (\ie the number of random projections) and the bandwidth $\s$ of the Gaussian kernel from which the random directions are drawn. 
The dimension $D$ was set to 2,048 for all experiments as there was negligible improvement in performance for larger values. 
$\s$ was chosen through cross-validation. % to be between 2.0 and 2.5 for SIFT descriptors and between 100 and 120 for color descriptors.
The choice of $\lambda$ (the regularization parameter of the GMP) has a significant impact on the final performance and was chosen 
by cross-validation from the set $\{10^1, 10^2, 10^3, 10^4, 10^5\}$.
We do not use spatial pyramids.

Results with the EMK are shown in Table~\ref{tab:emk_results}. 
We report results for the baseline EMK (sum-pooling \ie no mitigation of frequent descriptors) and the EMK with the proposed GMP.
A significant improvement in performance, between 3\% and 20\%, is achieved for all datasets when using GMP. 
This indicates that suppressing frequent descriptors is indeed beneficial when using EMKs.
On the fine-grained datasets, the improvements are particularly impressive -- 16\% on average.

\subsection{Results with the FV}
We now turn to the state-of-the-art FV representation~\cite{perronnin07fv,perronnin2010improving}.
To construct the FV we compute for each descriptor the gradient of the log-likelihood with respect 
to the parameters of a Gaussian Mixture Model (GMM) and pool the gradients. %Descriptors were assigned to visual words using hard assignment.
For the FV, increasing the number of Gaussians $G$ counteracts the negative effects of frequent descriptors
as fewer and fewer descriptors are assigned to the same Gaussian.
%of frequent descriptors by making it increasingly unlikely that different descriptors are pooled together
%(when a infinitely dimensional n. 
Therefore we expect GMP to have a smaller impact than for the EMK, 
particularly as $G$ increases.
By default we do not use spatial pyramids, but have included a discussion on its effect for the VOC-2007 dataset.

Experiments were conducted for FVs with $G$ set to either 16 or 256,
leading to 2,048-dim and 32,768-dim vectors respectively. 
Values of $G$ of 16 and 256 were chosen in order to have a comparable dimensionality to that of the EMK representation in the former case, 
and to have a state-of-the-art FV representation in the latter case. 
The value of the GMP regularization parameter $\lambda$ was once again chosen by cross-validation from the set $\{10^1, 10^2, 10^3, 10^4, 10^5\}$.\\[2mm]
\textbf{Power normalization baseline.}
Our baseline method uses power normalization, 
the state-of-the-art and post-hoc approach for improving the pooled FV representation~\cite{perronnin2010improving}. 
In the literature, the power is usually set to 0.5~\cite{perronnin2010improving,chatfield11devil,jegou2012aggregating,sanchez2013image}. 
Indeed, we found this value to be optimal for VOC-2007 for SIFT descriptors. 
However it has been shown, in the context of image retrieval, that a lower value
often can achieve significant performance gains~\cite{perronnin2010large}. 
We observed the same effect for classification. 
Therefore, %to obtain a strong baseline, 
we cross-validated the value of the power parameter.
We tested the following set of 8 values: $\{1.0, 0.7, 0.5, 0.4, 0.3, 0.2, 0.1, 0.0\}$.
Note that for a value of 0, we follow~\cite{perronnin2010large} and apply the power normalization only to non-zero entries.
Results with the best-performing power (\ie the value that led to the best results on the validation set) 
are denoted by sum+p in Table~\ref{tab:fv_results}.
The optimal power was determined on a per-descriptor and per-dataset basis. 
Hence, {\em our power baseline is a very strong one}.
For instance, for CUB-2011, performance with late fusion and $G=256$ increases from 25.4\% with the default $0.5$ value
to 29.7\% with the cross-validated power normalization.\\[2mm]
%Note that $\alpha=1$ corresponds to an unmodified FV with no power-normalization~\cite{perronnin07fv}.
%
\textbf{GMP \vs no power normalization.} 
Results are shown in Table~\ref{tab:fv_results}. 
Our GMP consistently performs significantly better -- 10\% better on average for late fusion and $G=256$ -- 
than when no power normalization is applied. 
The improvement is particularly impressive for several fine-grained datasets. 
For instance, for CUB-2011 GMP obtains a top-1 accuracy of 30.8\% compared to 13.2\% with sum-pooling.\\[2mm]
\begin{table}[!t]
\footnotesize
\setlength{\tabcolsep}{2pt}
\centering
\begin{tabular}{|l|cc|cc|cc|cc|cc|}
\hline
\multirow{2}{*}{Descriptor}& \multicolumn{2}{c|}{VOC-2007} & \multicolumn{2}{c|}{CUB-2010} & \multicolumn{2}{c|}{CUB-2011} & \multicolumn{2}{c|}{Pets} & \multicolumn{2}{c|}{Flowers}\\
\cline{2-11}
 & sum & GMP & sum & GMP & sum & GMP & sum & GMP & sum & GMP\\
\hline
SIFT & 40.0 & \textbf{46.0} & 2.9 & \textbf{6.5} & 5.0 & \textbf{10.6} & 21.7 & \textbf{35.6} & 41.3 & \textbf{52.2}\\
Color & 30.1 & \textbf{34.6} & 3.0 & \textbf{11.8} & 4.0 & \textbf{22.0} & 13.2 & \textbf{28.5} & 41.8 & \textbf{60.2}\\
\hline
Fusion & 43.1 & \textbf{49.5} & 3.7 & \textbf{13.5} & 6.0 & \textbf{24.8} & 22.8 & \textbf{42.9} & 55.8 & \textbf{69.5}\\
\hline
\end{tabular}
\vspace{0pt}
\caption{EMK results for SIFT descriptors, color descriptors, and late fusion of SIFT + color.
Results are shown for $D=2,048$-dim features for sum-pooling (sum) and for our GMP approach. }
\label{tab:emk_results}
\end{table}
\begin{table*}[!ht]
\footnotesize
\setlength{\tabcolsep}{1.5pt}
\centering
\begin{tabular}{|cl|cccc|cccc|cccc|cccc|cccc|}
\hline
\multicolumn{2}{|c|}{\multirow{2}{*}{Descriptor}}& \multicolumn{4}{c|}{VOC-2007} & \multicolumn{4}{c|}{CUB-2010} & \multicolumn{4}{c|}{CUB-2011} & \multicolumn{4}{c|}{Pets} & \multicolumn{4}{c|}{Flowers}\\
\cline{3-22}
& & sum & sum+p & GMP & GMP+p & sum & sum+p & GMP & GMP+p & sum & sum+p & GMP & GMP+p & sum & sum+p & GMP & GMP+p & sum & sum+p & GMP & GMP+p\\
\hline
\multirow{3}{*}{
\begin{sideways}
$G$=16
\end{sideways}} & \multicolumn{1}{|c|}{SIFT} & 49.1 & 51.6 & 52.8 & \textbf{53.4} & 4.1 & 6.2 & 6.4 & \textbf{6.9} & 7.9 & 11.1 & 11.5 & \textbf{12.7} & 29.4 & 32.1 & 35.1 & \textbf{35.7} & 58.3 & 62.8 & 63.8 & \textbf{65.3}\\
& \multicolumn{1}{|c|}{Color} & 40.2 & 43.8 & 45.3 & \textbf{45.8} & 4.9 & 8.7 & 12.5 & \textbf{13.1} & 7.2 & 16.8 & 21.6 & \textbf{22.8} & 22.5 & 28.6 & 32.5 & \textbf{33.5} & 55.3 & 65.6 & 65.9 & \textbf{67.2}\\
\cline{2-22}
& \multicolumn{1}{|c|}{Fusion} & 52.2 & 55.1 & 57.0 &\textbf{57.1} & 5.6 & 10.1 & 13.0 & \textbf{13.9} & 10.0 & 18.0 & 23.4 & \textbf{25.6} & 33.5 & 40.5 & 42.9 & \textbf{44.4} & 69.9 & 77.6 & 79.0 & \textbf{79.3}\\
\hline
\multirow{3}{*}{
\begin{sideways}
$G$=256
\end{sideways}} & \multicolumn{1}{|c|}{SIFT} & 52.6 & 57.8 & 58.1 & \textbf{58.9} & 5.3 & 8.1 & 7.7 & \textbf{9.6} & 10.2 & 16.3 & 16.4 & \textbf{17.0} & 38.1 & 47.1 & 47.9 & \textbf{49.2} & 67.7 & 73.0 & 72.8 & \textbf{73.3}\\
& \multicolumn{1}{|c|}{Color} & 39.4 & 49.5 & 50.0 & \textbf{50.4} & 4.6 & 13.0 & 14.2 & \textbf{14.6} & 9.0 & 27.4 & 27.0 & \textbf{29.3} & 23.6 & 41.1 & 41.6 & \textbf{43.0} & 63.9 & 74.0 & 72.8 & \textbf{75.1}\\
\cline{2-22}
& \multicolumn{1}{|c|}{Fusion} & 54.7 & 60.6 & \textbf{61.8} & 61.7 & 7.1 & 14.2 & 13.3 & \textbf{17.2} & 13.2 & 29.7 & 30.8 & \textbf{33.3} & 40.5 & 54.2 & 56.1 & \textbf{56.8} & 77.2 & 83.5 & 82.2 & \textbf{84.6}\\
\hline
\end{tabular}
\vspace{2pt}
\caption{FV results for SIFT descriptors, color descriptors, and late fusion of SIFT + color. 
Results are shown for $G=16$ and $G=256$ Gaussians 
for sum-pooling (sum), sum-pooling + power normalization (sum+p), our GMP approach (GMP), and GMP + power normalization (GMP+p).}
\label{tab:fv_results}
\end{table*}
\textbf{GMP \vs power normalization.} 
GMP always outperforms power normalization for all datasets for $G=16$.
The average improvement for late fusion is 2.8\%.
As expected, as $G$ increases to 256 GMP has less of an impact, 
but still outperforms power normalization by 0.4\% on average, with late fusion. 
Note that, on the Flowers dataset with late fusion and  $G=256$, 
we obtain 83.5\% and  82.2\% respectively for the power normalization and GMP.
These outperform the previous state-of-the-art (80.7\% \cite{angelova13}).
Also, on the Pets dataset with late fusion and $G=256$, 
GMP obtains top-1 accuracy of 56.1\%, 
compared to 54.2\% with power normalization -- an increase in performance of 1.9\%.
This is to our knowledge the best-reported result for this dataset, 
out-performing the previous state-of-the-art (54.3\% \cite{angelova13}). 
Therefore GMP achieves or exceeds the performance of the ad-hoc power normalization technique, 
while being more principled and more general. 
Note that GMP may also be combined with power normalization (GMP+p in Table~\ref{tab:fv_results}). 
This combination results in average improvements 
over power normalization of 3.8\% for $G=16$ and 2.5\% for $G=256$,
showing that GMP and power normalization are somewhat complementary.\\[2mm]
% as it is applicable to all image representations which aggregate local descriptors, including the EMK-based representation. We recall that applying the simple power heuristic to EMKs is not straight-forward, particularly for efficient evaluation of the kernel response.\\[2mm]
%
\textbf{Effect of spatial pyramids.}
We ran additional experiments on VOC-2007 to investigate the effect of our method when using Spatial Pyramids (SPs). 
We used a coarse pyramid and extracted 4 FVs per image: 
one FV for the whole image and one FV each for three horizontal strips 
corresponding to the top, middle and bottom regions of the image. 
With SPs, GMP again afforded improvements with respect to power normalization.
For instance, with late fusion and $G=256$, GMP obtains 62.0\% compared to 60.2\% for the power baseline --
a 1.8\% increase in performance.\\[2mm]
\textbf{Effect of the number of Gaussians $G$.} As expected, there is a consistent and significant positive impact on performance when $G$ is increased from 16 to 256. 
Our GMP approach is complementary to increasing $G$, as performance is generally improved when more Gaussians are used and GMP is applied. 
Furthermore, GMP is particularly attractive when low-dimensional FVs must be used.\\[2mm]
\textbf{FV \vs. EMK.} The baseline EMK results are quite poor in comparison with the baseline FV results. 
However, for CUB-2010, CUB-2011, and Pets, GMP improves the EMK performance to the point that EMK results with GMP are comparable to FV results 
with GMP when $G=16$ (with $G=16$, the FV and EMK representations are both 2,048-dimensional). 
In fact, for CUB-2011, EMK with GMP is superior to FV with GMP for $G=16$ (24.8\% \vs 23.4\%).

%% file: conclusions.tex
\section{Conclusions}\label{sec:conclusions}
We have proposed a principled and general method for pooling patch-level descriptors 
which equalizes the influence of frequent and rare descriptors, 
preserving discriminating information in the resulting aggregated representation. 
Our generalized max-pooling (GMP) is applicable to any encoding technique and 
can thus be seen as an extension of max pooling, 
which can only be applied to count-based representations such as BOV and its soft-coding
and hard-coding extensions. 
Extensive experiments on several public datasets show that GMP performs on par with, 
and sometimes significantly better than, heuristic alternatives.

We note that, in the same proceedings, J\'egou and Zisserman \cite{jegou2014triangulation} 
propose a democratic aggregation which bears some similarity to our GMP.
Especially, it involves re-weighting the patch encodings to balance 
the influence of descriptors.
One potential benefit of the GMP is that it is can be efficiently 
computed in the primal while the computation of the democratic aggregation can only be performed in the dual.
%which may be prohibitive for a large number of patches.
However, it remains to be seen how these two aggregation mechanisms compare in practice.

%% file: appendix.tex
\appendix

\section{GMP and Max-Pooling}
\label{sec:app}

In this appendix, we relate the GMP to max-pooling.
We denote by ${\cal E} = \{\p_1, \ldots, \p_N\}$ the set of descriptor encodings of a given image.
We assume that these encodings are drawn from a finite codebook of possible encodings, 
\ie $\p(x_i) \in \{q_1, \ldots, q_C\}$.
Note that the codewords $q_k$ might be binary or real-valued.
We denote by $Q$ the $D \times C$ codebook matrix of possible embeddings
where we recall that $D$ is the output encoding dimensionality.
We assume $Q = [q_1, \ldots, q_C]$ is orthonormal, \ie 
$Q^TQ = I_C$ where $I_C$ is the $C \times C$ identity matrix.
%The codebook might be different from one image to another.
For instance, in the case of the BOV with hard-coding,
%as the patch embeddings are drawn from a finite codebook $Q$.
$D=C$ and the $q_k$'s are binary with only the $k$-th 
entry equal to 1, so that $Q=I_C$.
We finally denote by $\pi_k$ the proportion of occurrences of $q_k$ in ${\cal E}$.

\vspace{3mm}

{\bf Theorem.}
$\p^{gmp}$ does not depend on the proportions $\pi_k$,
but only on the presence or absence of the $q_k$'s in ${\cal E}$.

\vspace{3mm}

{\bf Proof.}
We denote by $\Pi$ the $C \times C$ diagonal matrix that contains the values $\pi_1$, ..., $\pi_C$
on the diagonal. 
We rewrite $\Phi \1_M = Q \Pi \1_C$ and $\Phi \Phi^T = Q \Pi Q^T$.
The latter quantity is the SVD decomposition of $\Phi \Phi^T$ 
and therefore we have $(\Phi \Phi^T)^+ = Q \Pi^+ Q^T$.
Hence (\ref{eqn:primalsol}) becomes $\p^{gmp} = Q \Pi^+ Q^T Q \Pi \1_C = Q (\Pi^+ \Pi) \1_C$.
Since $\Pi$ is diagonal, its pseudo-inverse is diagonal and the values on the diagonal are equal to $1/\pi_k$ if
$\pi_k \neq 0$ and 0 if $\pi_k = 0$.
Therefore, $\Pi^+ \Pi$ is a diagonal matrix with element $k$ on the diagonal equal to 
$1$ if $\pi_k \neq 0$ and 0 otherwise.
Therefore we have
\begin{equation} \label{eqn:thm}
\p^{gmp} = \sum_{k: \pi_k \neq 0} q_k,
\end{equation}
which does not depend on the proportions $\pi_k$,
just on the presence or absence of the $q_k$'s in ${\cal E}$.

\vspace{3mm}

For the BOV hard-coding case, equation (\ref{eqn:thm}) shows that $\p^{gmp}$ is 
a binary representation where each dimension informs on the presence/absence of each codeword in the image.
This is {\em exactly the max-pooled representation}.
Therefore, our pooling mechanism can be understood as a {\em generalization of max-pooling}.

Note that there is no equivalence between the standard max-pooling and the GMP in the soft- or sparse-coding cases.
One benefit of the GMP however is that it is independent of a rotation of the encodings.
This is not the case of the standard max-pooling which operates on a per-dimension basis.

%We can also study the BOV in the regularized case.
%Again, assuming hard assignment, $\p(x_i)$ is binary with a single entry corresponding
%to the codeword index. 
%Therefore $\Phi \1$ corresponds to the (unnormalized) BOV histogram
%and $\Phi \Phi^T$ is a diagonal matrix with the BOV histogram on the diagonal.
%%Then $\p_i \p_i^T$ is a matrix with zeros everywhere and a single 1 on the diagonal.
%%Therefore, we have $\bar{\p}$ the standard BOV vector obtained by average pooling and $C$ is the diagonal
%%matrix with $\bar{\p}$ on the diagonal: $C = \mbox{diag}(\bar{\p})$.
%In such a case, equation (\ref{eqn:primalsolreg}) rewrites as
%%
%\begin{equation}
%\p^*_{\lbd} = \frac{\Phi \1}{\Phi \1 + \lbd},
%\end{equation}
%%
%where the previous division should be understood as a term-by-term operation.
%With $\lbd$ infinitely small, we again recover the standard max-pooling mechanism.

%We now study the case of a non-orthonormal basis.
%Since this is much more difficult, we restrict ourselves to the case where we have two entried in the codebook: $q_1$ and $q_2$.

\section*{Acknowledgments}
The authors wish to warmly thank H. J\'egou and T. Furon for
fruitful discussions as well as P. Torr, A. Vedaldi and A. Zisserman
for useful comments.
This work was done in the context of the Project Fire-ID, supported
by the Agence Nationale de la Recherche (ANR-12-CORD-0016).